\documentclass{article}

\usepackage{microtype}
\usepackage{graphicx}
\usepackage{subfigure}
\usepackage{booktabs} % for professional tables
\usepackage{amsthm, amssymb}
\usepackage{amsmath}
\usepackage{adjustbox}
\usepackage{enumitem}
\usepackage{multirow}
\usepackage{pifont}
\usepackage{color, colortbl}
\usepackage[dvipsnames]{xcolor}

\newtheorem{theorem}{Theorem}[section]

\newtheorem{lemma}[theorem]{Lemma}

\usepackage{hyperref}

\def\equationautorefname~#1\null{Eq.~(#1)\null}
\newcommand{\aref}[1]{\hyperref[#1]{Appendix~\ref{#1}}} % defined for sections in Appendix!

\newcommand{\mypm}[1]{\color{gray}{\scriptsize{$\pm$#1}}}
\usepackage[accepted]{icml2021}

\icmltitlerunning{Spiking Neural Networks Calibration}

\begin{document}

\twocolumn[
\icmltitle{A Free Lunch From ANN: \\ Towards Efficient, Accurate Spiking Neural Networks Calibration}



\icmlsetsymbol{equal}{*}

\begin{icmlauthorlist}
\icmlauthor{Yuhang Li}{equal,uestc,Y}
\icmlauthor{Shikuang Deng}{equal,uestc}
\icmlauthor{Xin Dong}{H}
\icmlauthor{Ruihao Gong}{st}
\icmlauthor{Shi Gu}{uestc}
\end{icmlauthorlist}

\icmlaffiliation{uestc}{University of Electronic Science and Technology of China}
\icmlaffiliation{Y}{Yale University}
\icmlaffiliation{H}{Harvard University}
\icmlaffiliation{st}{SenseTime Research}

\icmlcorrespondingauthor{Yuhang Li}{yuhang.li@yale.edu}
\icmlcorrespondingauthor{Shi Gu}{gus@uestc.edu.cn}

\icmlkeywords{Machine Learning, ICML}

\vskip 0.3in
]



\printAffiliationsAndNotice{\icmlEqualContribution} % otherwise use the standard text.

\begin{abstract}
Spiking Neural Network (SNN) has been recognized as one of the next generation of neural networks. Conventionally, SNN can be converted from a pre-trained ANN by only replacing the ReLU activation to spike activation while keeping the parameters intact. Perhaps surprisingly, in this work we show that a proper way to \textit{calibrate} the parameters during the conversion of ANN to SNN can bring significant improvements. We introduce SNN Calibration, a cheap but extraordinarily effective method by leveraging the knowledge within a pre-trained Artificial Neural Network (ANN). Starting by analyzing the conversion error and its propagation through layers theoretically, we propose the calibration algorithm that can correct the error layer-by-layer. 
The calibration only takes a handful number of training data and several minutes to finish. Moreover, our calibration algorithm can produce SNN with state-of-the-art architecture on the large-scale ImageNet dataset, including MobileNet and RegNet. 
Extensive experiments demonstrate the effectiveness and efficiency of our algorithm. For example, our advanced pipeline can increase up to 69\% top-1 accuracy when converting MobileNet on ImageNet compared to baselines. Codes are released at \href{https://github.com/yhhhli/SNN_Calibration}{a GitHub repo}.

\end{abstract}

\section{Introduction}
\label{Intro}

Spiking neural networks~(SNNs) are based on the spiking neural behavior in biological neurons~\cite{hodgkin1952quantitative,izhikevich2003simple}. Each neuron in SNNs elicits a spike when its accumulated membrane potential exceeds a threshold, otherwise, it would stay inactive in the current time step. Compared with ANNs, the activation values in SNNs are binarized (i.e., neuromorphic computing~\cite{roy2019nature}), thus resulting in an advantage of energy efficiency for SNNs. Existing works reveal that on specialized hardware, SNNs can save energy by orders of magnitude compared with ANNs ~\cite{roy2019scaling,DENG2020294}. Another vital attribute of SNN is its ability to make inferences in a spatial-temporal paradigm. Specifically, the forwarding pass in SNN is repeated for $T$ steps to get the final result, where the final result is the expectation of the ultimate layer's output across $T$ steps. This allows the flexibility of adjusting $T$ to balance between the latency and accuracy of SNNs for different application scenarios.  

\begin{table}[t]
\footnotesize
\centering
\begin{adjustbox}{max width=0.48\textwidth}
\begin{tabular}{lccc}
\toprule
\textbf{Features} & \textbf{Training} & \textbf{Conversion} & \textbf{Calibration} \\
\midrule
Accuracy($T\le100$) & High &  Low  & High \\
Scalability & Tiny & Large & Large \\
Training Speed & Slow & Fast & Fast \\
\# Required Data & Full-set & 1000 & 1000 \\
Inference Speed & Fast & Slow & Fast \\
\bottomrule
\end{tabular}
\end{adjustbox}
\caption{Features comparison between SNN direct training, ANN-SNN conversion and our SNN calibration.} \label{tab_compare}
\end{table}

Conventionally, there are two distinct routes to obtain a functional SNN: (1) training SNN from scratch \cite{shrestha2018slayer,kheradpisheh2018stdp}, and (2) converting a pretrained ANN to SNN \cite{cao2015spiking,Diehl2015Fast}. For training from scratch, it is hard to adopt gradient-based optimization methods because of the non-differentiability of the binary activation function in SNNs \cite{neftci2019surrogate}. 
Although several approaches like surrogate gradients~\cite{wu2018spatio,shrestha2018slayer} and synaptic plasticity~\cite{kheradpisheh2018stdp} are proposed to mitigate this problem, training SNN from scratch still lacks the scalability to obtain an effective SNN on the ImageNet dataset. Another notable problem is the tremendous resources required to complete the training process. The binary acceleration cannot be employed in GPU training since no CUDA instructions support this kind of computation. As a result, training an SNN may require $T\times$ more time than ANN training. 

Besides directly training SNNs from scratch, another family of approaches is converting a pretrained ANN into SNN ~\cite{diehl2016conversion,rueckauer2017conversion,Sengupta2018Going}. 
The conversion process demands less computation and memory than training from scratch. Although some progress in SNN conversion is made, such as threshold balancing~\cite{Diehl2015Fast,Sengupta2018Going}, weight normalization~\cite{rueckauer2017conversion}, and soft-reset mechanism~\cite{rueckauer2017conversion, han2020rmp}, all of them fail to convert ANN with BN layers in low latency time steps ($\le 256$), which may significantly increase the latency especially for resource-limited devices. We think the simple \textit{copy-paste} of parameters without any dedicated calibration on SNN will inevitably result in activation mismatch.

In this work, we aim to obtain an SNN in extremely low latency (less than 256 time steps) and in extremely low cost. We choose to utilize a pre-trained ANN and convert it to SNN. Unlike previous conversion work which simply \textbf{transplants the weights parameters} to the SNN, in this work we show that the \textbf{activation transplating} is much more important. In order to accomplish this, we propose SNN calibration, a new technology family by calibrating the parameters in SNN to match the activation after conversion, and thus significantly narrow the gap in activation distribution between the source ANN and calibrated SNN. We summarize the comparison between our calibration method and the existing conversion \& training methods in \autoref{tab_compare}. 
The novel contributions of the paper are threefold:

\begin{itemize}[leftmargin=*]
    \item We formulate the conversion equation and divide the conventional conversion error into flooring error and clipping error. And then we analyze the error propagation through layers. 
    \item We propose layer-wise calibration algorithm to adjust the network parameters including weights, bias, and initial potential to diminish the conversion error. To accommodate different user requirements, we provide Light Pipeline and Advanced Pipeline to balance accuracy and practical utility.
    \item We verify our algorithms on large-scale datasets like ImageNet~\cite{deng2009imagenet}. In addition to ResNets and VGG networks in previous work, we test a lightweight model MobileNet~\cite{Howard2017MobileNetsEC} and a large model RegNetX-4GF~\cite{radosavovic2020regnet} (79.4\% top-1 accuracy) for the first time in {ANN-to-SNN} conversion. Our method can increase up to 69\% accuracy in Spiking MobileNet conversion with 256 time steps. 
\end{itemize}

\section{Related Work}
%For an SNN and an ANN with the same architecture and parameters, \citet{cao2015spiking} first discover that neurons with high spike frequency in SNN correspond to higher activation values in ANN. Then, weight normalization and threshold balancing methods based on the ReLU function and the IF model \cite{liu2001spike,barbi2003stochastic} are proposed to convert simple ANN to SNN \cite{Diehl2015Fast,diehl2016conversion}. The threshold balancing mainly contains two steps: one is to copy the ANN's parameters to the SNN; other is to use the ANN layer's maximum activation as the threshold of the corresponding layer in the SNN. 

For training-based SNN, there are several supervised learning algorithms divided into (1) synaptic plasticity and (2) surrogate gradient. Synaptic plasticity methods are based on time-sensitivity and update the connection weight via the two neurons' firing time interval \cite{kheradpisheh2018stdp,iyer2020classifying,li2019research}. They are more suitable for the neuromorphic image \cite{amir2017low} or rate coding from static images. On the other hand, surrogate gradient (spiking-based backpropagation) methods use a soft relaxed function to replace the hard step function and train SNN like RNN \cite{wu2018spatio,shrestha2018slayer}. They suffer from the computationally expensive and slow during the training process on complex network architecture\cite{rathi2019enabling}.

Unlike training from scratch, ANN-to-SNN conversion methods, such as data-based normalization~\cite{Diehl2015Fast,rueckauer2016theory} or threshold balancing~\cite{Diehl2015Fast,diehl2016conversion}, adapt to more complex situations~\cite{tavanaei2019deep}.
%but need sizeable time steps to obtain reliable accuracy \cite{DENG2020294}.
% The threshold balancing is divided into two steps: (1) copy ANN parameters to SNN and change ReLU function to IF model; (2) find the maximum ANN activation as the layer's threshold.
The major bottleneck of these methods is how to balance accuracy and inference latency as they require more than 2k time steps to get accurate results. 
Recently, many methods have been proposed to reduce the conversion loss and simulation length. The soft-reset also called the reset-by-subtraction mechanism, is the most common technique to address the potential reset's information loss \cite{rueckauer2016theory,han2020deep}. Our IF neuron model also adopts this strategy. \citet{rueckauer2017conversion} suggest using percentile threshold, which avoids picking the outlier in the activation distribution.
% means the threshold is equal to $99.9\%$ largest activation instead of the maximum. 
%SPIKE-NORM~\cite{Sengupta2018Going} first converts the deep structure networks such as VGG-16 and ResNet-20 to SNN.
% More deep architecture (VGG-16 and ResNet-20) networks are first converted to SNN via the SPIKE-NORM algorithm~\cite{Sengupta2018Going}.
Spike-Norm~\cite{Sengupta2018Going} tests architectures like VGG-16 and ResNet-20. In this work, we further extend the source architecture to MobileNet and RegNet.
RMP~\cite{han2020rmp} and TSC~\cite{han2020deep} achieves near-to-origin accuracy by adjusting the threshold according to the input and output spike frequency.
% \citet{rathi2019enabling} attempt to use a hybrid training method. They only use 1/10 of the simulation length required for converted SNNs and directly train the SNNs on training set for less than 20 epochs to converge.
\citet{deng2021optimal} decompose the conversion loss into each layer and reduce it via shifting bias. Low latency converted SNN is an on-going research challenge since it still requires a considerable amount of simulation length. At the same time, most SNN conversion work does not address the BN layers in low latency settings.

\section{Preliminaries}
\label{sec_pre}
\textbf{Neuron Model for ANN.} 
Considering the $\ell$-th fully-connected layer or convolutional layer in the ANNs, its forwarding process can be formulated as,
\begin{equation}
    \mathbf{x}^{(\ell+1)} = h(\mathbf{z}^{(\ell)})=h(\mathbf{W}^{(\ell)}\mathbf{x}^{(\ell)}),  1\le\ell\le n, 
    \label{eq_ann}
\end{equation}
where $\mathbf{x}^{(\ell)}$, $\mathbf{W}^{(\ell)}$ denote the input activation and weight parameters in that layer respectively, and $h(\cdot)$ is the ReLU activation function. One can optionally train a bias parameter $\mathbf{b}^{(\ell)}$ and add it to pre-activation.

\textbf{Neuron Model for SNN.} 
Here we use the Integrate-and-Fire (IF) neuron model \cite{liu2001spike,barbi2003stochastic}. In specific, suppose at time step $t$ the spiking neurons in layer $\ell$ receive its binary input $\mathbf{s}^{(\ell)}(t)\in \{0, V_{th}^{(\ell-1)}\}$, the neuron will update its temporary membrane potential by,
\begin{equation}
    \mathbf{v}^{(\ell)}_{temp}(t+1) = \mathbf{v}^{(\ell)}(t) + \mathbf{W}^{(\ell)}\mathbf{s}^{(\ell)}(t),
    \label{eq_integrate}
\end{equation}
where $\mathbf{v}^{(\ell)}(t)$ denotes the membrane potential at time step $t$, and $\mathbf{v}^{(\ell)}_{temp}(t+1)$ denotes the intermediate variable that would be used to determine the update from $\mathbf{v}^{(\ell)}(t)$ to $\mathbf{v}^{(\ell)}(t+1)$.
If this temporary potential exceeds a pre-defined threshold $V_{th}^{(\ell)}$, it would produce a spike output $\mathbf{s}^{(\ell+1)}(t)$ with the value of $V_{th}^{(\ell)}$. Otherwise, it would release no spikes, i.e. $\mathbf{s}^{(\ell+1)}(t) = 0$. The membrane potential at the next time step $t+1$ would then be updated by \textit{soft-reset} mechanism, also known as \textit{reset-by-subtraction}. Mathematically, we describe the updating rule as 
\begin{equation}
    \mathbf{v}^{(\ell)}(t+1) = \mathbf{v}^{(\ell)}_{temp}(t+1) - \mathbf{s}^{(\ell+1)}(t), \label{eq_snn_neuron}
\end{equation}
\begin{equation}
    \mathbf{s}^{(\ell+1)}(t) = \begin{cases}
     V_{th}^{(\ell)} & \text{if }\mathbf{v}^{(\ell)}_{temp}(t+1) \ge V_{th}^{(\ell)} \\
    0 & \text{otherwise } 
    \end{cases}.
    \label{eq_if}
\end{equation}
Note that $V_{th}^{\ell}$ can be distinct in each layer. Thus, we cannot represent the spike in the whole network with binary signals. 
This problem can be avoided by utilizing a weight normalization technique to convert the $\{0, V_{th}^{(\ell-1)}\}$ spike to $\{0,1\}$ spike in every layers, given by:
\begin{equation}
    \mathbf{W}^{(\ell)} \leftarrow \frac{V_{th}^{(\ell-1)}}{V_{th}^{(\ell)}}, \ \ \ V_{th}^{(\ell)} \leftarrow 1. 
\end{equation} 
Recursively applying the above euqalization, we can use 0,1 spike to represent the intermediate activation for each layer. For the rest of the paper, we shall continue using the notation of $\{0, V_{th}^{(\ell-1)}\}$ spike for simplicity.

As for the input to the first layer and the output of the last layer, we do not employ any spiking mechanism. We use the first layer to direct encode the static image to temporal dynamic spikes, this can prevent the undesired information loss of the Poission encoding. For the last layer output, we only integrate the pre-synaptic input and does not firing any spikes. This is because the output can be either positive or negative, yet \autoref{eq_if} can only convert the ReLU activation. 

\textbf{Converting ANN to SNN }
Compared with ANN, SNN employs binary activation (i.e. spikes) at each layer. To compensate the loss in representation capacity, researchers introduce the time dimension to SNN by repeating the forwarding pass $T$ times to get final results. Ideally, the converted SNN is expected to have approximately the same input-output function mapping as the original ANN, i.e.,
\begin{equation}
    \mathbf{x}^{(\ell)} \approx \bar{\mathbf{s}}^{(\ell)}= \frac{1}{T}\sum_{t=0}^T\mathbf{s}^{(\ell)}(t).
    \label{approx_output}
\end{equation}
In practice, the above approximation only holds when $T$ grows to 1k or even higher. However, high $T$ would lead to large inference latency thus damage SNN's practical utility.

\begin{figure}[t]
    \centering
    \includegraphics[width=0.9\linewidth]{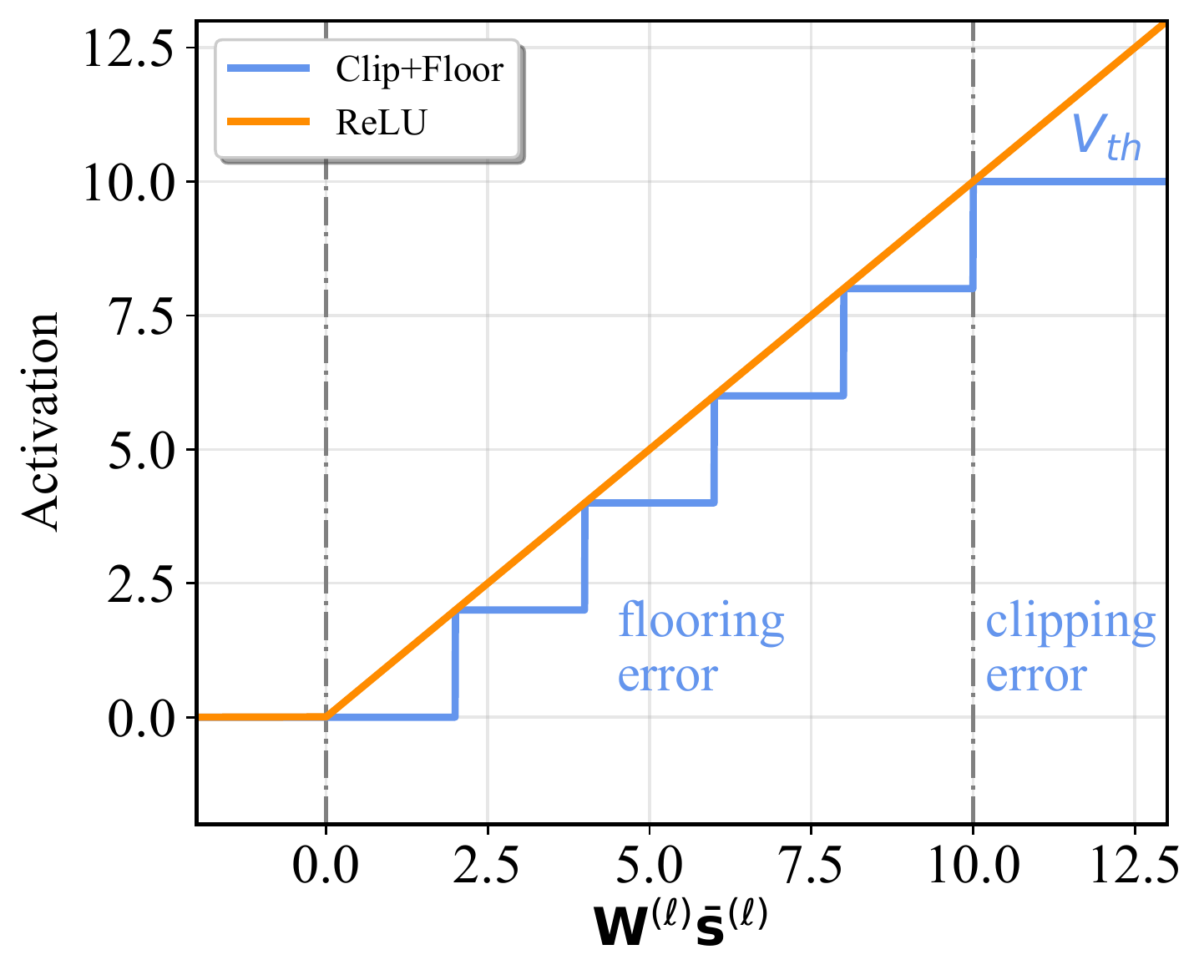}
    \caption{The conversion error between the ReLU activation used in ANN and the output spike in SNN ($V_{th}=10, T=5$) contains flooring error and clipping error.}
    \label{fig_clip_floor}
   \end{figure}

\section{Methodology}
\label{sec_method}
\subsection{Dividing the Conversion Loss}
We first use the derivation in \citet{deng2021optimal} to deduce the relationship between $\bar{\mathbf{s}}^{(\ell)}$ and $\bar{\mathbf{s}}^{(\ell+1)}$.
Suppose the initial membrane potential $\mathbf{v}^{(\ell)}(0) = \mathbf{0}$. Substitute \autoref{eq_integrate} into \autoref{eq_snn_neuron} and sum over $T$, then we get
\begin{equation}
   \mathbf{v}^{(\ell)}(T) = \mathbf{W}^{(\ell)}\left(\sum_{t=0}^{T}\mathbf{s}^{(\ell)}(t) \right) - \sum_{t=0}^{T}\mathbf{s}^{(\ell+1)}(t).
   \label{eq_T_total}
\end{equation}
Since at each time step, the output can be either $0$ or $V_{th}^{(\ell)}$, the accumulated output $\sum_{t=0}^{T}\mathbf{s}^{(\ell+1)}(t)$ can be written to $mV_{th}^{(\ell)}$ where $m \in \{0, 1, \dots, T\}$ denotes the total number of spikes. Note that we assume the terminal membrane potential $\mathbf{v}^{(\ell)}(T)$ lies within the range $[0, V_{th}^{(\ell)})$.
Therefore, according to \autoref{eq_T_total}, we have
\begin{equation}
    T\mathbf{W}^{(\ell)} \bar{\mathbf{s}}^{(\ell)} - V_{th}^{(\ell)} < mV_{th}^{(\ell)} \le T\mathbf{W}^{(\ell)}\bar{\mathbf{s}}^{(\ell)}.
\end{equation}
where $\bar{\mathbf{s}}^{(\ell)}$ is defined in \autoref{approx_output}. Then, we can use floor operation and clip operation to determine the $m$:
\begin{equation}
    m = \mathrm{clip}\left(\left\lfloor\frac{T}{V_{th}^{(\ell)}}\mathbf{W}^{(\ell)}\bar{\mathbf{s}}^{(\ell)}\right\rfloor, 0, T\right).
\end{equation}
Here the clip function sets the upper bound $T$ and lower bound $0$. Floor function $\lfloor x \rfloor$ returns the greatest integer that less than or equal to $x$. Given this formula, we can calculate the expected output spike:
\begin{align}
   \bar{\mathbf{s}}^{(\ell+1)} & = \mathrm{clipfloor}(\mathbf{W}^{(\ell)}\bar{\mathbf{s}}^{(\ell)}, T, V_{th}^{(\ell)}) \nonumber\\ 
   & = \frac{V_{th}^{(\ell)}}{T}\mathrm{clip}\left(\left\lfloor\frac{T}{V_{th}^{(\ell)}}\mathbf{W}^{(\ell)}\bar{\mathbf{s}}^{(\ell)}\right\rfloor, 0, T\right)
   \label{eq_clip_floor}
\end{align}
According to \autoref{eq_clip_floor}, the conversion loss (difference between $\mathbf{x}^{(\ell+1)}$ and $\bar{\mathbf{s}}^{(\ell+1)}$) comes from two aspects, namely the \emph{flooring error} and the \emph{clipping error}. 

In \autoref{fig_clip_floor}, we further indicate that $V_{th}^{(\ell)}$ is crucial for conversion loss because it affects both the flooring and the clipping errors. Increasing $V_{th}^{(\ell)}$ leads to lower clipping error but higher flooring error. % \xin{delete: For the simulation length $T$, increasing it will only decrease the flooring error.} 
Previous work \cite{Diehl2015Fast,diehl2016conversion} sets $V_{th}^{(\ell)}$ to the maximum pre-activations across samples in ANN to eliminate the clipping error. However, the maximum pre-activations are usually outliers. Given this insight, the outliers may tremendously increase the flooring error. As a result, they have to use a very large $T$ (for example, 2000) to decrease the flooring error. 

\begin{figure}[t]
 \centering
 \includegraphics[width=0.9\linewidth]{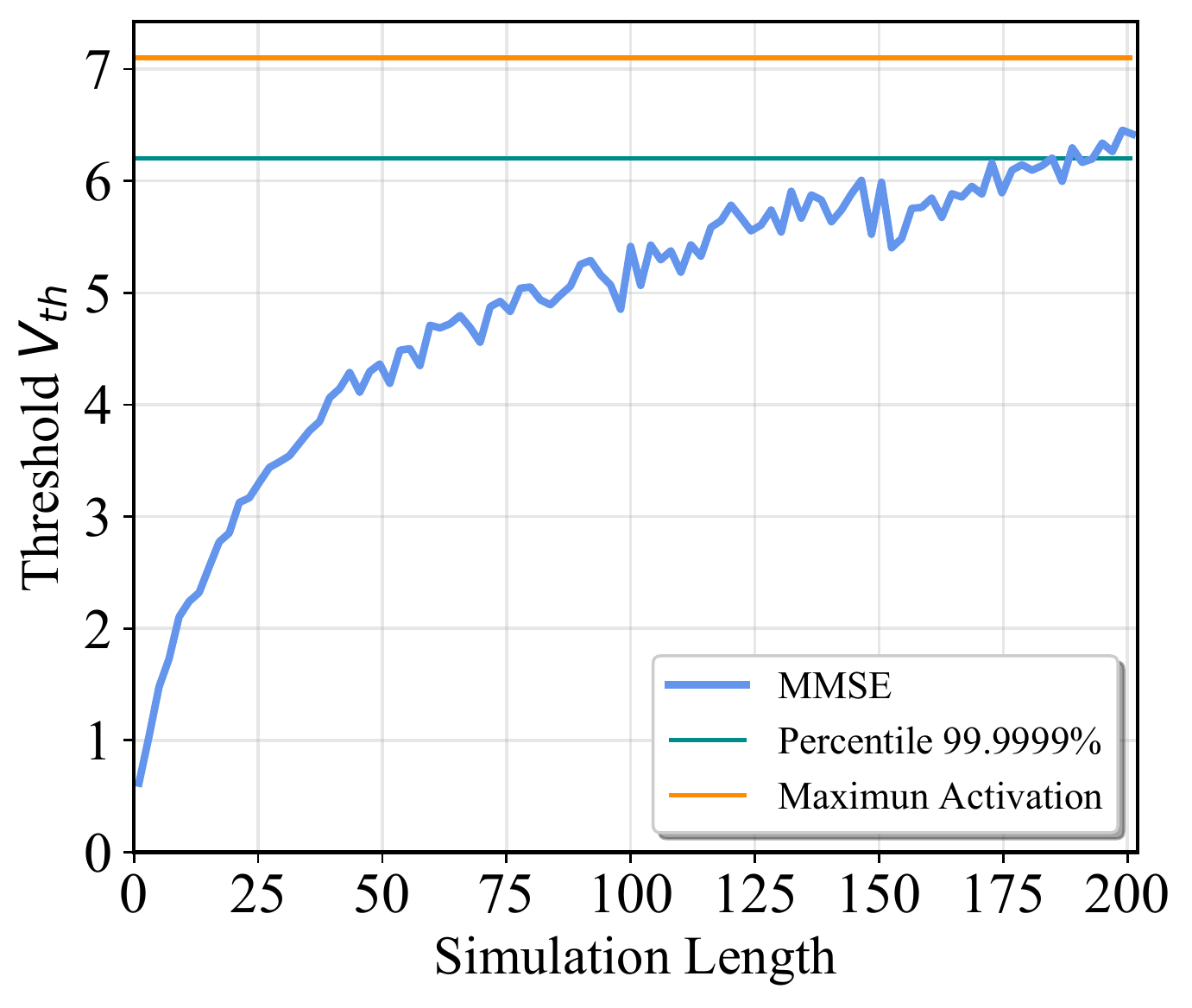}
 \caption{Comparison of the threshold determined by three different approaches. Our MMSE threshold will vary with time steps.}
 \label{fig_mse}
\end{figure}

\subsection{Adaptive Threshold by MMSE}
In an effort to better balance flooring error and the clipping error, we use Minimization of Mean Squared Error (MMSE) to obtain the threshold $V_{th}^{(\ell)}$ under different simulation length $T$. 
Here we adopt the similar layer-wise optimization problem in~\cite{deng2021optimal}, which is formulated by
\begin{equation}
    \min_{V_{th}} \left(\mathrm{clipfloor}(\mathbf{x}^{(\ell+1)}, T, V_{th}^{(\ell)}) - \mathrm{ReLU}({\mathbf{x}}^{(\ell+1)})\right)^2
\end{equation}
Note that the above problem is not guaranteed to be convex, and there is no closed-form solution to this minimization. We hereby sample several batches of training images and use grid search to determine the final result of $V_{th}^{(\ell)}$. Specifically, we linearly sample $N$ grids between $[0, \max({\mathbf{x}}^{(\ell+1)})]$, and find the grid that has lowest MSE. We set $N=100$ and find this option is precise enough to obtain a good solution.
\autoref{fig_mse} shows the dynamics of our proposed method. It is worthwhile to note that $V_{th}^{(\ell)}$ does not monotonically increase along with $T$, because the flooring error may be decreased by slightly increment the threshold. We can further apply MMSE threshold channel-wisely to further decrease the MSE error, as did in~\cite{kim2019spiking}.
% A further step to reduce the MSE in conversion is to use a fine-grained threshold, i.e., apply MMSE channel-wisely and then normalize this threshold into weights kernel as did in \citet{kim2019spiking}.

\subsection{Layer-wise Calibration}
\label{sec_calibration}
% With dynamic threshold tuning, we can get the optimal threshold. However, the conversion error still remain and have a great impact because of the low robustness.
%\xin{delete: Adaptive threshold only minimizes the conversion error if the parameters in ANN are simply copied to SNN.}
Besides adaptive threshold, we further reduce the conversion error by calibrating the parameters of SNN.
We first analyze how conversion errors accumulate through layers, 
and then present a set of layer-wise algorithms to calibrate different types of SNN parameters, including bias, weights and initial membrane potential. 

As aforementioned, the output layer of our SNN only accumulate the pre-synaptic inputs through time, that is to say, the final output of SNN is arithmetic sum of output from each time step $\sum^T_{t=1}\mathbf{W}^{(n)}\mathbf{s}^{(n)}(t)$.
%employ the same module as in the ANN rather than the Integrate-and-Fire module in order to take care of the negative output. 
Note that this modification doesn't introduce additional overhead because network output won't participate in further operation.
With this modified last layer, our object becomes to minimize the difference between the input to the last layer, i.e. $\bar{\mathbf{s}}^{(n)}$ and $\mathbf{x}^{(n)}$.
%Note that we do not apply IF neuron model for the last output layer in the network because the network output cannot always be positive like the ReLU activation. As a result, some negative output may never fire any spikes. 
%Meanwhile, the . So, disabling the IF neuron model in last layer does not hurt the overall efficiency.
%Instead, in the last layer we will only integrate the membrane potential and not fire any spikes (\autoref{eq_integrate}). Therefore, if the expected input spikes in the last layer $\bar{\mathbf{s}}^{(n)}$ equal to the input activation of ANN $\mathbf{x}^{(n)}$, we have $\mathbf{W}^{(n)}\bar{\mathbf{s}}^{(n)} = \mathbf{W}^{(n)}\mathbf{x}^{(n)}$ and apply $\mathrm{Softmax}(\cdot)$ can get the same classification results. 
% \xin{If my understand is correct, you just do not convert the last layer. If so, please simplify this paragraph.} 
%Then our task becomes to minimize the difference between $\bar{\mathbf{s}}^{(n)}$ and $\mathbf{x}^{(n)}$. 

\begin{lemma}
\label{lemma1}
Denote the Frobenius norm as $||\cdot||$, the conversion error in the last layer in given by
\begin{equation}
    ||\mathbf{e}^{(n)}||=||\mathbf{x}^{(n)} - \bar{\mathbf{s}}^{(n)}|| \le ||\sum_{\ell=1}^{n} Err^{(\ell)} \prod_{k=\ell}^{n}\mathbf{W}^{(k)}||,
    \label{eq_final_err}
\end{equation}
where $Err^{(\ell)}=\mathrm{clipfloor}(\bar{\mathbf{s}}^{(\ell)})-\mathrm{ReLU}(\bar{\mathbf{s}}^{(\ell)})$.
\end{lemma}

We provide the detailed derivation of above lemma in \aref{sec_append_A}. The above lemma indicates that the conversion errors in former layers have a cumulative effect on the subsequent layers. 
%For example, $Err^{(1)}$ will propagate to layer 2 till layer $n$.
In addition, the conversion error in the ultimate layer is upper bounded by weighted linear combination of layer-wise error. Based on this observation, we develop a set of greedy layer-wise calibration algorithm to correct the conversion error in each layer progressively.

% Based on different types of the network parameters, we introduce three different calibration algorithms: \emph{(1)} \emph{Bias Calibration (BC)} which adjusts bias $\mathbf{b}$, \emph{(2)} \emph{Potential Calibration (PC)} which adjusts the initial membrane potential $\mathbf{v}(0)$, and \emph{(3)} \emph{Weights Calibration (WC)} which adjusts weights $\mathbf{W}$. 
% \textbf{Light and Advanced Pipelines }
% SNN is known for the flexibility to make tradeoff between latency and accuracy by varying $T$. 
% However, to maintain high accuracy with relatively small $T$, we cannot simply adopt the weight and bias from the pre-trained model with high $T$ (ref to the existing works of conversion).
We introduce Light Pipeline and Advanced Pipeline, which can be chose by users according to their memory and computation budgets for layer-wise calibration in practice.
The light pipeline achieves fast calibration with less memory and computation consumption by only adjusting bias in SNNs. With a little effort, the light pipeline can outperform state-of-the-art methods by a large margin. 
We also propose an Advanced Pipeline that achieves best results by calibrating the weights as well as the initial membrane potential in a fine-grained way.

\textbf{Light Pipeline }
% Bias Calibration can reduce the conversion error by adjusting the bias parameters based on the expectation of $Err$. 
% In Light Pipeline, we only calibrate the bias parameters. 
Light Pipeline only contains Bias Calibration (BC).
In order to calibrate the bias parameters, we first define a reduced mean function:
\begin{equation}
    \mu_c(\mathbf{x})= \frac{1}{wh}\sum_{i=1}^{w}\sum_{j=1}^{h} \mathbf{x}_{c,i,j}
\end{equation}
where $w, h$ are the width and height of the feature-map, and $\mu_c(\mathbf{x})$ computes the spatial mean of the feature-map in each channel $c$.
We notice that the spatial mean of conversion error $\mathbf{e}^{(\ell)} = \mathbf{x}^{(\ell)} - \bar{\mathbf{s}}^{(\ell)}$ can be written by
\begin{equation}
   \mu_c(\mathbf{x}^{(\ell)}) = \mu_c(\bar{\mathbf{s}}^{(\ell)}) + \mu_c(\mathbf{e}^{(\ell)}). \label{eq_bc}
\end{equation}
%where $\mathbb{E}_{\mathcal{P}}$ is the expectation computed over data distribution $\mathcal{P}(\mathbf{x^{(1)}})$ and $\mathbb{E}_{c}$ is the expectation over output channel dimension.
To ensure the mean output of SNN is equal to the mean output of ANN, we can add the expected conversion error into the bias term as $\mathbf{b}^{(\ell)}_c\leftarrow\mathbf{b}^{(\ell)}_c+\mu_c(\mathbf{e}^{(\ell+1)})$. In practice, we only sample one batch training images and compute the reduced mean to calibrate the bias. 
% In BC, we compute the mean of $Err^{(\ell)}$ and directly add it to the bias term in SNN model. Bias Calibration is fast and only requires a single batch to compute the $Err^{(\ell)}$. 

\textbf{Advanced Pipeline }
Calibrating the bias only corrects partial error. We need a more fine-grained calibration method. To this end, we propose advanced pipeline which consists of Potential Calibration (PC) and Weights Calibration (WC).

Now consider a non-zero initial membrane potential $\mathbf{v}^{(\ell)}(0)$, we can rewrite \autoref{eq_clip_floor} to 
\begin{equation}
      \tilde{\bar{\mathbf{s}}}^{(\ell+1)} = \frac{V_{th}^{(\ell)}}{T}\mathrm{clip}\left(\left\lfloor\frac{T}{V_{th}^{(\ell)}}\mathbf{W}^{(\ell)}\bar{\mathbf{s}}^{(\ell)}+\frac{\mathbf{v}^{(\ell)}(0)}{V_{th}^{(\ell)}}\right\rfloor, 0, T\right),
\end{equation}
where $\tilde{\bar{\mathbf{s}}}^{(\ell+1)}$ is the calibrated expected output with non-zero initialization of membrane potential.
To obtain a fast calibration for the initial membrane potential, here we make an approximation that:
\begin{align}
    \tilde{\bar{\mathbf{s}}}^{(\ell+1)} & \approx \frac{V_{th}^{(\ell)}}{T}\mathrm{clip}\left(\left\lfloor\frac{T}{V_{th}^{(\ell)}}\mathbf{W}^{(\ell)}\bar{\mathbf{s}}^{(\ell)}\right\rfloor, 0, T\right)+\frac{\mathbf{v}^{(\ell)}(0)}{T} \nonumber \\
    & = \bar{\mathbf{s}}^{(\ell+1)} + \mathbf{v}^{(\ell)}(0)/T
\end{align} 
Similar to BC, $\mathbf{v}^{(\ell)}(0)/T$ can correct the output distribution of SNN. We can directly set $\mathbf{v}^{(\ell)}(0)$ to $T\times \mathbf{e}^{(\ell+1)}$ to calibrate the initial potential. 
%As a result, PC directly determines the initial membrane potential by $T\times \mathbb{E}[\mathbf{x}^{(\ell+1)}-\bar{\mathbf{s}}^{(\ell+1)}]$.
Note that Potential Calibration does not need to compute spatial mean.

In advanced pipeline, we also introduce Weights Calibration to correct the conversion error in each layer. In WC, the formulation is given by:
\begin{align}
   \min_{\mathbf{W}^{(\ell)}}  ||\mathbf{e}^{(\ell+1)}||^2 \label{eq_wc}.
\end{align}
Here we optimize the whole weights tensor in SNN layer-by-layer and can reduce the conversion error even further. For practical implementation, we will first store input samples in ANN $\mathbf{x}^{(\ell)}$ and input spikes samples in SNN of each time step $\mathbf{s}^{(\ell)}(t)$ and then compute the expected spike input by $\bar{\mathbf{s}}^{(\ell)}=\sum_{t=1}^T\mathbf{s}^{(\ell)}(t)$. To further compute the gradient of $\mathrm{clipfloor}$ function, we apply the StraightThrough Estimator~\cite{bengio2013ste} of the floor operation, i.e. 
\begin{equation}
    \frac{\partial \lfloor x \rfloor}{\partial x} = 1.
\end{equation} 
To this end, we can use regular training methods like stochastic gradient descent for calibrating the weights.
% Compared to the regular SNN training method, WC is computationally efficient, because the expected output $\bar{\mathbf{s}}^{(\ell-1)}$ can be pre-computed and we only have to do one-time convolution rather than $T$ times.
When conducting calibration for weights, the optimization process is very efficient compared to other direct training methods. This is because we first store the expected input from previous layers, and we do not have to perform $T$ times convolution like direct training methods. The major bottleneck of WC is storing the input of SNN. For example, if we set $T=1024$, then we will do 1024 times forwarding pass for one batch and accumulate them to get the final expected results.

\begin{algorithm}[t]
\caption{Overall Algorithms}
\label{alg:1}
\begin{algorithmic}
    \INPUT Pretrained ANN; simulation length $T$
    \STATE Fold BN Layers into Conv Layers (cf. \autoref{eq_bn_fold})
    \STATE Replace AvgPooling Layers to depthwise Conv Layers
    \FOR{all $i=1,2,\dots, N$-th layers in the ANN}
        \STATE Collect input data $\mathbf{x}^{(i)}$, output data $\mathbf{x}^{(i+1)}$ in one batch
        \STATE Get MMSE threshold $V_{th}^{(i)}$ using grid search
        \STATE Get SNN output $\bar{\mathbf{s}}^{(i+1)}$
        \STATE Compute Error term $\mathbf{e}^{(i+1)}=\mathbf{x}^{(i+1)}-\bar{\mathbf{s}}^{(i+1)}$
        \IF{\textit{Light Pipeline}}
            \STATE Calibrate bias term $\mathbf{b}^{(i)}\leftarrow\mathbf{b}^{(i)}+\mu(\mathbf{e}^{(i+1)})$
        \ELSE
            \STATE Calibrate Potential $\mathbf{v}^{(i)}(0) \leftarrow T\times\mathbf{e}^{(i+1)}$ 
            \STATE Optimize weights to minimize $||\mathbf{e}^{(\ell+1)}||^2$  via stochastic gradient descent
        \ENDIF
    \ENDFOR
\OUTPUT Converted SNN model
\end{algorithmic}
\end{algorithm}

\subsection{Average Pooling Layers}
%\textbf{Input and Output Layers }
%For the input layers, we do not convert images to binary spike information because this operation will damage the conversion performance \cite{rueckauer2017conversion} and requires time to generate the binary signals. We also do not output any spikes in the last layer but only accumulate the membrane potential as mentioned in the former section.
% \cite{lu2020exploring} why add this ref? 
%First, this does not increase the additional cost or destroy SNN characteristics. Second, only using the positive part of the output may significantly reduce network performance for complex tasks \cite{kim2019spiking}. 

%Most SNN works do not include Max Pooling layers since choosing the maximum activation in spatial-temporal dimension is difficult. 
Most SNN works do not include the Max Pooling layers since finding the maximum activation neuron ahead of time is impractical, i.e. we cannot determine the maximum neuron $\bar{\mathbf{s}}$ when we only observe $\mathbf{s}(1)$. 
Therefore, they use Average Pooling Layers to downsample the feature-maps and do not convert them in SNN. However, we argue that Average Pooling Layers will produce non-binary information. For example, a $2\times2$ kernel AvgPool layer can output 4 possible values $[0, 0.25, 0.5, 1]$ with spike inputs. 
To make SNN run on corresponding hardware, we convert the AvgPool layer by treating the AvgPool layer as a convolutional layer with specific values. So we can convert the AvgPool layer just like other convolutional layers. More details are included in Appendix.

\subsection{Converting BN layers}
There is no corresponding module in SNN for Batch Normalization (BN) layers. \citet{rueckauer2017conversion} propose to absorb the BN parameters to the weight and bias, which can be represented by:
\begin{equation}
    \mathbf{W}\leftarrow \mathbf{W}\frac{\gamma}{\sigma}, \ \ \ \mathbf{b}\leftarrow \beta + (\mathbf{b}-\mu)\frac{\gamma}{\sigma}, \label{eq_bn_fold}
\end{equation}
where $\mu, \sigma$ are the running mean and standard deviation, and $\gamma, \beta$ are the transformation parameters in the BN layer. 

\section{Experiments}

To demonstrate the effectiveness and the efficiency of the proposed algorithm, we conduct experiments on CIFAR~\cite{cifardataset} and ImageNet~\cite{deng2009imagenet} datasets with extremely low simulation length (say $T\le 256$). In \autoref{sec_ab_study}, we study the impact of the approximations and design choices made in~\autoref{sec_method}. 
In \autoref{sec_sota_comp}, we compare our methods to other methods. 

\subsection{Implementation Details}
For all ANN with BN layers, we fold the BN layer before conversion. We do not convert input images to binary spikes because generating binary spikes requires time and degrades the accuracy. We also do not convert network output to spikes as explained in~\autoref{sec_calibration}.
To correct the bias and membrane potential, we sample one batch of unlabeled data (128 training images). To estimate the MMSE threshold and calibrate weights, we use 1024 training images. 
In our experiments, we apply the bias shift as described in \citet{deng2021optimal}. 
We use Stochastic Gradient Descent with 0.9 momentum to optimize weights in WC, followed by a cosine learning rate decay~\cite{loshchilov2016sgdr}. The learning rate for WC is set to $10^{-5}$, and no L2 regularization is imposed. We optimize the weights in each layer with 5000 iterations. We will analyze the time and space complexity of our algorithm in the next section. Note that the training details of ANN are included in the Appendix.

\subsection{Ablation Study}
In this section, we verify the design choices of our proposed adaptive threshold and layer-wise calibration. In all ablation experiments, we test \textit{VGG-16 and ResNet-20}~\cite{Sengupta2018Going,han2020rmp} \textit{on CIFAR100}.
We also conduct variance studies by running 5 times with different random seeds and report the mean and standard deviation of the (top1) accuracy on the validation set.%, calculated using 5 runs with different initial seeds. 

\label{sec_ab_study}

\begin{table}[t]
\footnotesize
    \centering
    \begin{adjustbox}{max width=0.48\textwidth}
    \begin{tabular}{lcccc}
    \toprule
    \multirow{2}{6em}{\textbf{Method}} & \multicolumn{2}{c}{\textbf{VGG-16 (77.89)}} & \multicolumn{2}{c}{\textbf{ResNet-20 (77.16)}} \\
    \cmidrule(l{2pt}r{2pt}){2-3}\cmidrule(l{2pt}r{2pt}){4-5}
    & T=16 & T=32 & T=16 & T=32 \\
    \midrule
    Maximum Act & 2.38\mypm{0.17} & 5.01\mypm{1.23}& 25.67\mypm{4.67}& 51.27\mypm{3.82}\\
    Percentile 99.9\% & 3.73\mypm{0.38} & 42.11\mypm{0.37} & 55.00\mypm{0.47} & 71.94\mypm{0.19} \\
    MMSE (Ours) & \textbf{19.42\mypm{0.81}} & 43.53\mypm{0.72} & 58.38\mypm{0.62} & 72.13\mypm{0.18} \\
    MMSE* (Ours) & 17.50\mypm{2.46} & \textbf{47.40\mypm{2.71}} & \textbf{63.55\mypm{0.60}} & \textbf{73.57\mypm{0.06}} \\
    \bottomrule
    \end{tabular}
    \end{adjustbox}
    \caption{(Top-1) Accuracy comparison on different threshold determination methods for ANN with BN layers. * denotes channel-wise threshold.} \label{tab_mmse_thresh}
\end{table}

\begin{table}[t]
\footnotesize
    \centering
    \begin{adjustbox}{max width=0.48\textwidth}
    \begin{tabular}{lcccc}
    \toprule
    \multirow{2}{6em}{\textbf{Method}} & \multicolumn{2}{c}{\textbf{VGG-16 (77.89)}} & \multicolumn{2}{c}{\textbf{ResNet-20 (77.16)}} \\
    \cmidrule(l{2pt}r{2pt}){2-3}\cmidrule(l{2pt}r{2pt}){4-5}
    & T=16 & T=32 & T=16 & T=32 \\
    \midrule
    MaxAct + BC& 24.61\mypm{2.19} & 32.60\mypm{2.61} & 55.17\mypm{5.06} & 68.78\mypm{2.48} \\
    Percentile + BC & 40.56\mypm{1.29} & 66.87\mypm{0.78} & 69.60\mypm{0.14}& 75.26\mypm{0.20} \\
    MMSE + BC & 44.95\mypm{0.52} & 67.61\mypm{0.72} & 70.78\mypm{0.15} & 75.53\mypm{0.15}\\
    MMSE* + BC & \textbf{52.96\mypm{1.91}} & \textbf{69.19\mypm{0.75}} & \textbf{72.33\mypm{0.13}} & \textbf{75.94\mypm{0.23}} \\
    \bottomrule
    \end{tabular}
    \end{adjustbox}
    \caption{Light pipeline combines BC with different thresholds for ANN with BN layers and consistently improves accuracy. * denotes channel-wise threshold.} \label{tab_bias_corr}
\end{table}

\textbf{Effect of MMSE Threshold in Conversion}
%Then 
We study the effect of choosing different threshold $V_{th}$. In \autoref{tab_mmse_thresh}, we show that maximum activation has the lowest effect because of the under-fire problem in the initial stage. Our MMSE threshold achieves better results than maximum activation~\cite{Diehl2015Fast} and percentile~\cite{rueckauer2017conversion} threshold when $T =16$. As an example, our method is 15.7\% higher in accuracy than percentile when converting VGG-16. In ResNet-20, better threshold can significantly improve the accuracy of conversion. 
%Although percentile threshold has similar results with our MMSE on Spiking VGG-16 ($T=32$), we argue that our MMSE is a $T$-dependent method and obtain a similar results with percentile threshold in $T=32$.
% Although percentile threshold has similar results to our layer-wise MMSE when the simulation time is 32, \GS{we argue that our MMSE is a $T$-dependent method with a sustained privilege over the percentile strategy (see Appendix XXX)}.
% In our experiment, the layer-wise MMSE is better than the percentile if we continue to increase $T$. 
%MMSE also has lower variance in experiments compared to Percentile threshold.
Finally, we apply the channel-wise MMSE threshold and further boost the accuracy from 43.5\% to 47.4\% in VGG-16 and from 72.1\% to 73.5\% in ResNet-20.

\textbf{Light Pipeline: Combining Bias Calibration}
Next, we verify the effect of the proposed Bias Calibration by applying it to different threshold methods. Results are summarized in \autoref{tab_bias_corr}, where we can find BC can \textit{consistently improve the accuracy of converted SNN by simply tuning the bias parameters.} For example, BC can boost 22\% accuracy in VGG-16 using percentile threshold when $T=16$. On our MMSE threshold, the Bias Calibration can increase up to 35\% accuracy. We should emphasize that BC is cheap and only requires tiny memory space to store the bias term for different $T$. Therefore our light pipeline is flexible to make trade-off between accuracy and latency.

%----------------------Big Table Here -----------------------
\newcommand{\cmark}{\color{ForestGreen}\ding{51}}%
\newcommand{\xmark}{\color{Red}\ding{55}}%

\begin{table*}[h]
    \footnotesize
        \centering
        \begin{adjustbox}{max width=\textwidth}
        \begin{tabular}{lcccccccc}
        \toprule
        \textbf{Method} & \textbf{Use BN} & \textbf{Convert AP} & \textbf{ANN Acc.} & \textbf{$T=32$} & \textbf{$T=64$} & \textbf{$T=128$} &\textbf{ $T=256$} & \textbf{ $T\ge 2048$} \\
        \midrule
        \multicolumn{9}{c}{\textbf{ResNet-34~\cite{He2016Deep}    ImageNet}} \\
        \midrule
        Spike-Norm~\cite{Sengupta2018Going} & \xmark & \xmark & 70.69 & - &- &- &- & 65.47 \\ 
        Hybrid Train~\cite{rathi2019enabling} & \xmark & \xmark & 70.20 & - &- &- & 61.48 & 65.10 \\
        RMP~\cite{han2020rmp} & \xmark & \xmark & 70.64 & - & - & - & 55.65 & 69.89 \\
        TSC~\cite{han2020deep} & \xmark & \xmark & 70.64 & - & - & - & 55.65 & 69.93 \\
        Opt.~\cite{deng2021optimal}* & \xmark & \cmark & 70.95 & 33.01 & 59.52 & 67.54 & 70.06 & 70.98 \\
        \textbf{Ours (Light Pipeline)} & \xmark & \cmark & 70.95 & \textbf{62.34} & \textbf{68.38} & \textbf{70.15} & \textbf{70.75} & 70.97 \\
        Opt.~\cite{deng2021optimal}* & \cmark & \cmark & 75.66 & 0.09 & 0.12 & 3.19 & 47.11 & 75.08 \\
        \textbf{Ours (Light Pipeline)} & \cmark & \cmark & 75.66 & \textbf{50.21} & \textbf{63.66} & \textbf{68.89} & \textbf{72.12} &  75.44 \\
        \textbf{Ours (Advanced Pipeline)} & \cmark & \cmark & 75.66 & \textbf{64.54} & \textbf{71.12} & \textbf{73.45} & \textbf{74.61} & 75.45 \\
        \midrule
        \multicolumn{9}{c}{\textbf{VGG-16~\cite{simonyan2014very} ImageNet}} \\
        \midrule
        Spike-Norm~\cite{Sengupta2018Going} & \xmark & \xmark & 70.52 & - &- &- &- & 69.96 \\ 
        Hybrid Train~\cite{rathi2019enabling} & \xmark & \xmark & 69.35 & - &- &- & 62.73 & 65.19 \\
        RMP~\cite{han2020rmp} & \xmark & \xmark & 73.49 & - & - & - & 48.32 & 73.09 \\
        TSC~\cite{han2020deep} & \xmark & \xmark & 73.49 & - & - & - & 69.71 & 73.46 \\
        Opt.~\cite{deng2021optimal}* & \xmark & \cmark & 72.40 & 54.92 & 66.51 & 69.94 & 71.35 & 72.09 \\
        \textbf{Ours (Light Pipeline)} & \xmark & \cmark & 72.40 & \textbf{69.30} & \textbf{71.12} & \textbf{71.85} & \textbf{72.20} & 72.29 \\
        Opt.~\cite{deng2021optimal}* & \cmark & \cmark & 75.36 & 0.114 & 0.118 & 0.122 & 1.81 & 73.88 \\
        \textbf{Ours (Light Pipeline)} & \cmark & \cmark & 75.36 & \textbf{24.88} & \textbf{56.77} & \textbf{70.49} & \textbf{73.66} & 75.15 \\
        \textbf{Ours (Advanced Pipeline)} & \cmark & \cmark & 75.36 & \textbf{63.64} & \textbf{70.69}  & \textbf{73.32}  & \textbf{74.23} & 75.32 \\
        \midrule
        \multicolumn{9}{c}{\textbf{MobileNet~\cite{Howard2017MobileNetsEC} ImageNet}} \\ 
        \midrule
        Opt.~\cite{deng2021optimal}* & \cmark & \cmark & 73.40 & 0.110 & 0.104  & 0.100 & 0.964 & 68.21 \\
        \textbf{Ours (Light Pipeline)} & \cmark & \cmark & 73.40 & 0.254 & \textbf{12.62} & \textbf{53.91}  & \textbf{65.86} & \textbf{72.19}\\
        \textbf{Ours (Advanced Pipeline)} & \cmark & \cmark & 73.40 & \textbf{37.43} & \textbf{56.26} & \textbf{65.40} & \textbf{69.02} & \textbf{72.38} \\ 
         \midrule
        \multicolumn{9}{c}{\textbf{RegNetX-4GF~\cite{radosavovic2020regnet} ImageNet}} \\ 
        \midrule
        Opt.~\cite{deng2021optimal}* & \cmark & \cmark & \textbf{80.02} & 0.218 & 3.542 & 48.60 & 71.22 & 78.33 \\
        \textbf{Ours (Light Pipeline)} & \cmark & \cmark & \textbf{80.02} & \textbf{35.63} & \textbf{65.28} & \textbf{74.37}  & \textbf{77.33} & \textbf{79.15}\\
        \textbf{Ours (Advanced Pipeline)} & \cmark & \cmark & \textbf{80.02} & \textbf{55.70} & \textbf{70.96} & \textbf{75.78} & \textbf{77.50} & \textbf{79.21} \\
        \bottomrule
        \end{tabular}
        \end{adjustbox}
        \caption{Comparison of our algorithm with other existing SNN conversion works. \textit{Use BN} means use BN layers to optimize ANN, \textit{Convert AP} means use depthwise convolutional layers to replace Average Pooling layers. * denotes self-implementation results.} \label{tab_sota_compare}
\end{table*}

% -----------------------------------------------------------

\begin{table}[t]
\footnotesize
    \centering
    \begin{adjustbox}{max width=0.48\textwidth}
    \begin{tabular}{lcccc}
    \toprule
    \multirow{2}{6em}{\textbf{Method}} & \multicolumn{2}{c}{\textbf{VGG-16 (77.89)}} & \multicolumn{2}{c}{\textbf{ResNet-20 (77.16)}} \\
    \cmidrule(l{2pt}r{2pt}){2-3}\cmidrule(l{2pt}r{2pt}){4-5}
    & T=16 & T=32 & T=16 & T=32 \\
    \midrule
    MMSE + PC & 59.52\mypm{1.06} & 70.62\mypm{0.47} & 73.37\mypm{0.28} & 76.21\mypm{0.06}\\
    MMSE* + PC & 65.29\mypm{0.86} & 73.53\mypm{0.27} & \textbf{74.22\mypm{0.25}} & \textbf{76.68\mypm{0.12}} \\
    \midrule
    MMSE + PC + WC & 65.02\mypm{0.33}& 73.51\mypm{0.23} & 73.36\mypm{0.28} & 76.32 \mypm{0.13}\\
    MMSE* + PC + WC & \textbf{67.14\mypm{0.88}} & \textbf{74.52\mypm{0.36}} & 74.02\mypm{0.20} & 76.52\mypm{0.16}\\
    \bottomrule
    \end{tabular}
    \end{adjustbox}
    \caption{Advanced pipeline use Potential and Weights Calibration to optimize SNN. * denotes channel-wise threshold.} \label{tab_heavy_pipe}
\end{table}
\textbf{Advanced Pipeline: Potential and Weights Calibration }
Our advanced pipeline contains Potential and Weights Calibration that will alter the ANN's parameters to adapt better in spiking configuration. We validate the effect of them on our MMSE threshold mode in \autoref{tab_heavy_pipe}. We can find that Potential Calibration can substantially improve the accuracy of SNN. As an example, the MMSE + BC on VGG-16 only has 44.95\% accuracy. However, with PC, we can uplift the accuracy to 59.52\%. We also find WC is slightly more stable since the variance of the results is lower. 


\subsection{Comparison to Previous Work}
\label{sec_sota_comp}

In this section, we compare our proposed algorithm with other existing work. We first test ImageNet models\footnote{We include the comparison on CIFAR dataset in \aref{sec_cifar_results}. }. Here we choose the widely adopted ResNet-34~\cite{He2016Deep} and VGG-16~\cite{simonyan2014very} in the existing literature. Note that we test ANNs both with and without BN layers. We additionally verify our algorithm on \textbf{MobileNet}~\cite{Howard2017MobileNetsEC}. To our best knowledge, this is the first work that studies Spiking MobileNet conversion.

Results can be found in \autoref{tab_sota_compare}. For both ResNet-34 and VGG-16 without BN, our light pipeline is within 1\% accuracy loss when $T=128$. On models with BN layers, our method can substantially improve the conversion loss. In particular, the light pipeline can improve 50.1\%, and the advanced pipeline can improve 64.4\% accuracy in ResNet-34 with BN Conversion when $T=32$. Baseline methods still produce a large accuracy gap even on VGG-16 without BN layers and AvgPool Conversion. While our light pipeline reaches 73.66 (less than 2\% accuracy drop) when $T=256$. 
The superiority of our algorithm is also reflected in Spiking MobileNet conversion, where the baseline method \cite{deng2021optimal} crashed when $T\le 256$. To reach acceptable accuracy of Spiking MobileNet, the baseline method has to increase $T$ to 2048. However, our advanced pipeline can achieve higher accuracy while reducing 8$\times$ simulation length (69.02 when $T=256$). Finally, we test our algorithm on a large ANN, RegNetX-4GF~\cite{radosavovic2020regnet} which achieves 79.4\% top-1 accuracy. Our light pipeline reaches 73\% accuracy when $T=128$ and our advanced pipeline reaches 75.8\% accuracy when $T=128$.

\subsection{Complexity Study}
\textbf{Time Complexity }
During run-time, our converted SNN will not produce additional inference time. However, converting SNN using light or advanced pipeline may require time and computing resources. The time needed for each calibration is described in the table below. All experiments were tested on a single NVIDIA GTX 1080TI with 5 runs. In \autoref{tab_convert_time}, we can see that the Bias Calibration and Potential Calibration only takes limited time on CIFAR100 and ImageNet. Using Weights Calibration is much expensive than the other two methods. For example, calibrate a MobileNet on ImageNet may take 30 minutes using the advanced pipeline. We should emphasize that our advanced pipeline is still much cheaper than Hybrid Train~\cite{rathi2019enabling}, which requires 20 epochs of end-to-end training (hundreds of GPU hours).

\textbf{Space Complexity }
In this section, we report the memory requirements for each calibration algorithm. Since our method will calibrate a new set of parameters for different $T$, therefore it is necessary to study the model size if we want to deploy SNNs under different $T$. Specifically, calibrating the bias of ResNet-34 on ImageNet only requires 0.3653MB memory. However, calibrating the weights and potential requires 83.25MB and 18.76MB, respectively. 
Thus, our proposed light pipeline is both computational and memory cheap and is optimal for flexible SNN conversion. In contrast, the advanced pipeline (PC and WC) requires much more memory space. One may optionally only apply PC to lower down the memory footprint of ResNet-34. Interestingly, some tiny structures like MobileNet share less weights memory (12.21MB) but higher activation memory (19.81MB).

\begin{table}[t]
    \footnotesize
        \centering
        \begin{adjustbox}{max width=0.48\textwidth}
        \begin{tabular}{lccc}
        \toprule
        \textbf{Model} & \textbf{BC}  & \textbf{PC} & \textbf{WC} \\
        \midrule
        VGG-16 & 0.098$\pm$0.003 & 0.106$\pm$0.017 & 4.70$\pm$0.037 \\
        MobileNet & 2.29$\pm$0.005 & 2.19$\pm$0.01 & 27.6$\pm$1.62 \\ 
        \bottomrule
        \end{tabular}
        \end{adjustbox}
        \caption{Conversion time (minutes) for different calibration algorithm. We set $T=64$ and test VGG-16 on CIFAR100 and MobileNet on ImageNet.} \label{tab_convert_time}
\end{table}

\begin{table}[t]
    \footnotesize
        \centering
        \begin{adjustbox}{max width=0.48\textwidth}
        \begin{tabular}{lrrrr}
        \toprule
        \textbf{\#Samples} & 32  & 64 & 128 & 256 \\
        \midrule
        VGG-16 & 64.50$\pm$1.19 & 65.12$\pm$1.02 & 66.04$\pm$0.72 & 66.20$\pm$0.58\\
        ResNet-20 & 75.35$\pm$0.16 & 75.41$\pm$0.18 & 75.43$\pm$0.09 & 75.41$\pm$0.08\\ 
        \bottomrule
        \end{tabular}
        \end{adjustbox}
        \caption{Comparison of the accuracy using different number of data samples for bias correction.} \label{tab_num_sample}
\end{table}

\textbf{Data Sample Complexity}
We study the robustness of the our algorithm by increasing the size of calibration dataset. Here we test Bias Calibration in~\autoref{tab_num_sample} on ResNet-20 and VGG-16 ($T=32$). By increasing the number of samples for calibration, the accuracy will also increase. However, we can see that in ResNet-20 the effect of samples is trivial. While in VGG-16, increasing the number of samples from 32 to 256 can increase 1.7\% mean accuracy. We also find that more samples lead to a stable calibration result. Therefore, we recommend using at least 128 images for calibration. In our experiments, the same trend is also observed in other calibration algorithms.

\subsection{Efficiency and Sparsity}
\begin{figure}[t]
    \centering
    \includegraphics[width=0.45\textwidth]{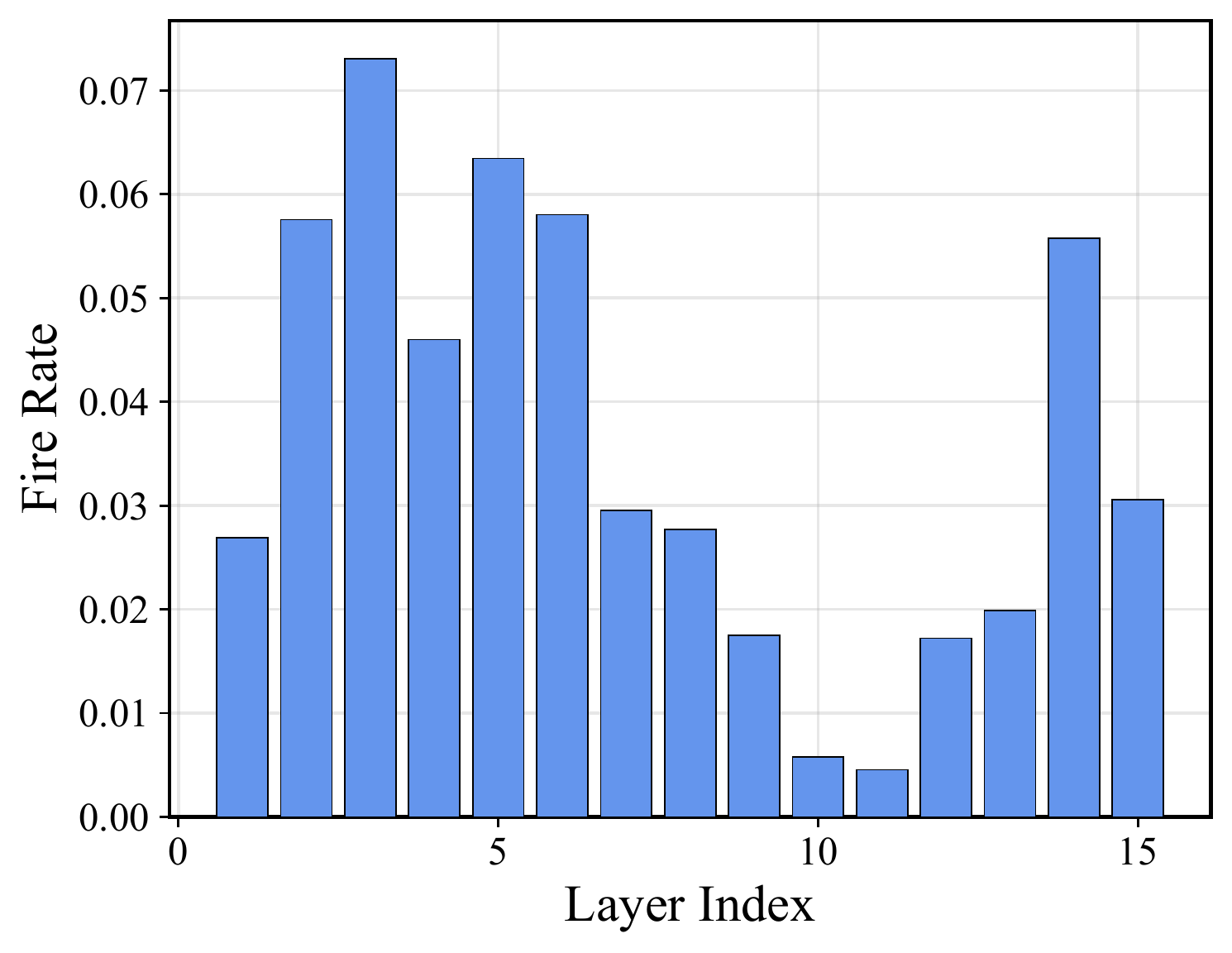}
    \caption{Firing rate visualization of VGG-16.}
    \label{fig_fire_rate}
\end{figure}

In this section we visualize the sparsity of our calibrated SNN. We choose the spike VGG-16 on the ImageNet dataset, with $T=64$. We only leverage light pipeline (bias calibration) and record the mean firing ratio across the whole validation dataset, which also corresponds to sparsity of the activation. The firing ratio is demonstrated in \autoref{fig_fire_rate}, where we can find the maximum firing ratio is under 0.08, and the minimum firing ratio can be 0.025. To quantitaively compute the energy saving, we use the energy-estimation euqation in~\citet{rathi2020diet}. For addition, we measure it by 0.9$J$ per operation; for multiplication, we measure it by 4.6$J$ per operation. On the event-driven neuromorphic hardware, a non-firing neuron will not cost any energy. Based on this rule, our calibrated spiking VGG-16 only costs 69.36\% energy of ANN's consumption.

\section*{Conclusion}
In this work, we analyze the composition of conversion error and its cumulative effect. To reduce the gap between ANN activation and SNN activation, we propose adaptive threshold to determine the threshold in different time steps. We also introduce the layer-wise calibration, which significantly improves the performance of SNN compared with other \emph{simple-copy} methods. Layer-wise calibration is easy to use and only requires a few training images. 
Our method establishes new state-of-the-art performance for SNN conversion. It can successfully convert challenging architectures like MobileNet and RegNetX-4GF with a low latency (less than 256 time steps) for the first time. Even when converting the ANN with Batch Normalization layers, our method can preserve high classification accuracy. 

\section*{Acknowledgement}
This work is supported by NSFC 61876032. We greatly thank annoymous reviewers for their kind suggestions to this work.

\bibliography{bibfile}

\begin{thebibliography}{39}
\providecommand{\natexlab}[1]{#1}
\providecommand{\url}[1]{\texttt{#1}}
\expandafter\ifx\csname urlstyle\endcsname\relax
  \providecommand{\doi}[1]{doi: #1}\else
  \providecommand{\doi}{doi: \begingroup \urlstyle{rm}\Url}\fi

\bibitem[Amir et~al.(2017)Amir, Taba, Berg, Melano, McKinstry, Di~Nolfo, Nayak,
  Andreopoulos, Garreau, Mendoza, et~al.]{amir2017low}
Amir, A., Taba, B., Berg, D., Melano, T., McKinstry, J., Di~Nolfo, C., Nayak,
  T., Andreopoulos, A., Garreau, G., Mendoza, M., et~al.
\newblock A low power, fully event-based gesture recognition system.
\newblock In \emph{Proceedings of the IEEE Conference on Computer Vision and
  Pattern Recognition}, pp.\  7243--7252, 2017.

\bibitem[Barbi et~al.(2003)Barbi, Chillemi, Di~Garbo, and
  Reale]{barbi2003stochastic}
Barbi, M., Chillemi, S., Di~Garbo, A., and Reale, L.
\newblock Stochastic resonance in a sinusoidally forced lif model with noisy
  threshold.
\newblock \emph{Biosystems}, 71\penalty0 (1-2):\penalty0 23--28, 2003.

\bibitem[Bengio et~al.(2013)Bengio, L{\'e}onard, and Courville]{bengio2013ste}
Bengio, Y., L{\'e}onard, N., and Courville, A.
\newblock Estimating or propagating gradients through stochastic neurons for
  conditional computation.
\newblock \emph{arXiv preprint arXiv:1308.3432}, 2013.

\bibitem[Cao et~al.(2015)Cao, Chen, and Khosla]{cao2015spiking}
Cao, Y., Chen, Y., and Khosla, D.
\newblock Spiking deep convolutional neural networks for energy-efficient
  object recognition.
\newblock \emph{International Journal of Computer Vision}, 113\penalty0
  (1):\penalty0 54$--$66, 2015.

\bibitem[Cubuk et~al.(2019)Cubuk, Zoph, Mane, Vasudevan, and
  Le]{cubuk2019autoaugment}
Cubuk, E.~D., Zoph, B., Mane, D., Vasudevan, V., and Le, Q.~V.
\newblock Autoaugment: Learning augmentation strategies from data.
\newblock In \emph{Proceedings of the IEEE/CVF Conference on Computer Vision
  and Pattern Recognition}, pp.\  113--123, 2019.

\bibitem[Deng et~al.(2009)Deng, Dong, Socher, Li, Li, and
  Fei-Fei]{deng2009imagenet}
Deng, J., Dong, W., Socher, R., Li, L.-J., Li, K., and Fei-Fei, L.
\newblock Imagenet: A large-scale hierarchical image database.
\newblock In \emph{2009 IEEE conference on computer vision and pattern
  recognition}, pp.\  248--255. Ieee, 2009.

\bibitem[Deng et~al.(2020)Deng, Wu, Hu, Liang, Ding, Li, Zhao, Li, and
  Xie]{DENG2020294}
Deng, L., Wu, Y., Hu, X., Liang, L., Ding, Y., Li, G., Zhao, G., Li, P., and
  Xie, Y.
\newblock Rethinking the performance comparison between snns and anns.
\newblock \emph{Neural Networks}, 121:\penalty0 294 -- 307, 2020.

\bibitem[Deng \& Gu(2021)Deng and Gu]{deng2021optimal}
Deng, S. and Gu, S.
\newblock Optimal conversion of conventional artificial neural networks to
  spiking neural networks.
\newblock In \emph{International Conference on Learning Representations}, 2021.
\newblock URL \url{https://openreview.net/forum?id=FZ1oTwcXchK}.

\bibitem[DeVries \& Taylor(2017)DeVries and Taylor]{devries2017cutout}
DeVries, T. and Taylor, G.~W.
\newblock Improved regularization of convolutional neural networks with cutout.
\newblock \emph{arXiv preprint arXiv:1708.04552}, 2017.

\bibitem[Diehl et~al.(2015)Diehl, Neil, Binas, Cook, and Liu]{Diehl2015Fast}
Diehl, P.~U., Neil, D., Binas, J., Cook, M., and Liu, S.~C.
\newblock Fast-classifying, high-accuracy spiking deep networks through weight
  and threshold balancing.
\newblock In \emph{Neural Networks (IJCNN), 2015 International Joint Conference
  on}, 2015.

\bibitem[Diehl et~al.(2016)Diehl, Zarrella, Cassidy, Pedroni, and
  Neftci]{diehl2016conversion}
Diehl, P.~U., Zarrella, G., Cassidy, A., Pedroni, B.~U., and Neftci, E.
\newblock Conversion of artificial recurrent neural networks to spiking neural
  networks for low-power neuromorphic hardware.
\newblock In \emph{2016 IEEE International Conference on Rebooting Computing
  (ICRC)}, pp.\  1--8. IEEE, 2016.

\bibitem[Han \& Roy(2020)Han and Roy]{han2020deep}
Han, B. and Roy, K.
\newblock Deep spiking neural network: Energy efficiency through time based
  coding.
\newblock In \emph{European Conference on Computer Vision}, 2020.

\bibitem[Han et~al.(2020)Han, Srinivasan, and Roy]{han2020rmp}
Han, B., Srinivasan, G., and Roy, K.
\newblock Rmp-snn: Residual membrane potential neuron for enabling deeper
  high-accuracy and low-latency spiking neural network.
\newblock In \emph{Proceedings of the IEEE/CVF Conference on Computer Vision
  and Pattern Recognition}, pp.\  13558--13567, 2020.

\bibitem[He et~al.(2016)He, Zhang, Ren, and Jian]{He2016Deep}
He, K., Zhang, X., Ren, S., and Jian, S.
\newblock Deep residual learning for image recognition.
\newblock In \emph{2016 IEEE Conference on Computer Vision and Pattern
  Recognition}, 2016.

\bibitem[He et~al.(2019)He, Zhang, Zhang, Zhang, Xie, and Li]{he2019bag}
He, T., Zhang, Z., Zhang, H., Zhang, Z., Xie, J., and Li, M.
\newblock Bag of tricks for image classification with convolutional neural
  networks.
\newblock In \emph{Proceedings of the IEEE/CVF Conference on Computer Vision
  and Pattern Recognition}, pp.\  558--567, 2019.

\bibitem[Hodgkin \& Huxley(1952)Hodgkin and Huxley]{hodgkin1952quantitative}
Hodgkin, A.~L. and Huxley, A.~F.
\newblock A quantitative description of membrane current and its application to
  conduction and excitation in nerve.
\newblock \emph{The Journal of physiology}, 117\penalty0 (4):\penalty0
  500--544, 1952.

\bibitem[Howard et~al.(2017)Howard, Zhu, Chen, Kalenichenko, Wang, Weyand,
  Andreetto, and Adam]{Howard2017MobileNetsEC}
Howard, A.~G., Zhu, M., Chen, B., Kalenichenko, D., Wang, W., Weyand, T.,
  Andreetto, M., and Adam, H.
\newblock Mobilenets: Efficient convolutional neural networks for mobile vision
  applications.
\newblock \emph{arXiv preprint arXiv:1704.04861}, 2017.

\bibitem[Iyer \& Chua(2020)Iyer and Chua]{iyer2020classifying}
Iyer, L.~R. and Chua, Y.
\newblock Classifying neuromorphic datasets with tempotron and spike timing
  dependent plasticity.
\newblock In \emph{2020 International Joint Conference on Neural Networks
  (IJCNN)}, pp.\  1--8. IEEE, 2020.

\bibitem[Izhikevich(2003)]{izhikevich2003simple}
Izhikevich, E.~M.
\newblock Simple model of spiking neurons.
\newblock \emph{IEEE Transactions on neural networks}, 14\penalty0
  (6):\penalty0 1569--1572, 2003.

\bibitem[Kheradpisheh et~al.(2018)Kheradpisheh, Ganjtabesh, Thorpe, and
  Masquelier]{kheradpisheh2018stdp}
Kheradpisheh, S.~R., Ganjtabesh, M., Thorpe, S.~J., and Masquelier, T.
\newblock Stdp-based spiking deep convolutional neural networks for object
  recognition.
\newblock \emph{Neural Networks}, 99:\penalty0 56--67, 2018.

\bibitem[Kim et~al.(2019)Kim, Park, Na, and Yoon]{kim2019spiking}
Kim, S., Park, S., Na, B., and Yoon, S.
\newblock Spiking-yolo: Spiking neural network for energy-efficient object
  detection.
\newblock \emph{arXiv preprint arXiv:1903.06530}, 2019.

\bibitem[Krizhevsky et~al.()Krizhevsky, Nair, and Hinton]{cifardataset}
Krizhevsky, A., Nair, V., and Hinton, G.
\newblock Cifar-10 (canadian institute for advanced research).
\newblock URL \url{http://www.cs.toronto.edu/~kriz/cifar.html}.

\bibitem[LI \& LI(2019)LI and LI]{li2019research}
LI, S.-L. and LI, J.-P.
\newblock Research on learning algorithm of spiking neural network.
\newblock In \emph{2019 16th International Computer Conference on Wavelet
  Active Media Technology and Information Processing}, pp.\  45--48. IEEE,
  2019.

\bibitem[Liu \& Wang(2001)Liu and Wang]{liu2001spike}
Liu, Y.-H. and Wang, X.-J.
\newblock Spike-frequency adaptation of a generalized leaky integrate-and-fire
  model neuron.
\newblock \emph{Journal of computational neuroscience}, 10\penalty0
  (1):\penalty0 25--45, 2001.

\bibitem[Loshchilov \& Hutter(2016)Loshchilov and Hutter]{loshchilov2016sgdr}
Loshchilov, I. and Hutter, F.
\newblock Sgdr: Stochastic gradient descent with warm restarts.
\newblock \emph{arXiv preprint arXiv:1608.03983}, 2016.

\bibitem[Neftci et~al.(2019)Neftci, Mostafa, and Zenke]{neftci2019surrogate}
Neftci, E.~O., Mostafa, H., and Zenke, F.
\newblock Surrogate gradient learning in spiking neural networks: Bringing the
  power of gradient-based optimization to spiking neural networks.
\newblock \emph{IEEE Signal Processing Magazine}, 36\penalty0 (6):\penalty0
  51--63, 2019.

\bibitem[Radosavovic et~al.(2020)Radosavovic, Kosaraju, Girshick, He, and
  Doll{\'a}r]{radosavovic2020regnet}
Radosavovic, I., Kosaraju, R.~P., Girshick, R., He, K., and Doll{\'a}r, P.
\newblock Designing network design spaces.
\newblock In \emph{Proceedings of the IEEE/CVF Conference on Computer Vision
  and Pattern Recognition}, pp.\  10428--10436, 2020.

\bibitem[Rathi \& Roy(2020)Rathi and Roy]{rathi2020diet}
Rathi, N. and Roy, K.
\newblock Diet-snn: Direct input encoding with leakage and threshold
  optimization in deep spiking neural networks.
\newblock \emph{arXiv preprint arXiv:2008.03658}, 2020.

\bibitem[Rathi et~al.(2019)Rathi, Srinivasan, Panda, and
  Roy]{rathi2019enabling}
Rathi, N., Srinivasan, G., Panda, P., and Roy, K.
\newblock Enabling deep spiking neural networks with hybrid conversion and
  spike timing dependent backpropagation.
\newblock In \emph{International Conference on Learning Representations}, 2019.

\bibitem[Roy et~al.(2019{\natexlab{a}})Roy, Chakraborty, and
  Roy]{roy2019scaling}
Roy, D., Chakraborty, I., and Roy, K.
\newblock Scaling deep spiking neural networks with binary stochastic
  activations.
\newblock In \emph{2019 IEEE International Conference on Cognitive Computing
  (ICCC)}, pp.\  50--58. IEEE, 2019{\natexlab{a}}.

\bibitem[Roy et~al.(2019{\natexlab{b}})Roy, Jaiswal, and Panda]{roy2019nature}
Roy, K., Jaiswal, A., and Panda, P.
\newblock Towards spike-based machine intelligence with neuromorphic computing.
\newblock \emph{Nature}, 575\penalty0 (7784):\penalty0 607--617,
  2019{\natexlab{b}}.

\bibitem[Rueckauer et~al.(2016)Rueckauer, Lungu, Hu, and
  Pfeiffer]{rueckauer2016theory}
Rueckauer, B., Lungu, I.-A., Hu, Y., and Pfeiffer, M.
\newblock Theory and tools for the conversion of analog to spiking
  convolutional neural networks.
\newblock \emph{arXiv: Statistics/Machine Learning}, \penalty0
  (1612.04052):\penalty0 0--0, 2016.

\bibitem[Rueckauer et~al.(2017)Rueckauer, Lungu, Hu, Pfeiffer, and
  Liu]{rueckauer2017conversion}
Rueckauer, B., Lungu, I.-A., Hu, Y., Pfeiffer, M., and Liu, S.-C.
\newblock Conversion of continuous-valued deep networks to efficient
  event-driven networks for image classification.
\newblock \emph{Frontiers in neuroscience}, 11:\penalty0 682, 2017.

\bibitem[Sengupta et~al.(2018)Sengupta, Ye, Wang, Liu, and
  Roy]{Sengupta2018Going}
Sengupta, A., Ye, Y., Wang, R., Liu, C., and Roy, K.
\newblock Going deeper in spiking neural networks: Vgg and residual
  architectures.
\newblock \emph{Frontiers in Neuroence}, 13, 2018.

\bibitem[Shrestha \& Orchard(2018)Shrestha and Orchard]{shrestha2018slayer}
Shrestha, S.~B. and Orchard, G.
\newblock Slayer: Spike layer error reassignment in time.
\newblock In \emph{Advances in Neural Information Processing Systems}, pp.\
  1412--1421, 2018.

\bibitem[Simonyan \& Zisserman(2014)Simonyan and Zisserman]{simonyan2014very}
Simonyan, K. and Zisserman, A.
\newblock Very deep convolutional networks for large-scale image recognition.
\newblock \emph{arXiv preprint arXiv:1409.1556}, 2014.

\bibitem[Szegedy et~al.(2016)Szegedy, Vanhoucke, Ioffe, Shlens, and
  Wojna]{szegedy2016labelsmooth}
Szegedy, C., Vanhoucke, V., Ioffe, S., Shlens, J., and Wojna, Z.
\newblock Rethinking the inception architecture for computer vision.
\newblock In \emph{Proceedings of the IEEE conference on computer vision and
  pattern recognition}, pp.\  2818--2826, 2016.

\bibitem[Tavanaei et~al.(2019)Tavanaei, Ghodrati, Kheradpisheh, Masquelier, and
  Maida]{tavanaei2019deep}
Tavanaei, A., Ghodrati, M., Kheradpisheh, S.~R., Masquelier, T., and Maida, A.
\newblock Deep learning in spiking neural networks.
\newblock \emph{Neural Networks}, 111:\penalty0 47--63, 2019.

\bibitem[Wu et~al.(2018)Wu, Deng, Li, Zhu, and Shi]{wu2018spatio}
Wu, Y., Deng, L., Li, G., Zhu, J., and Shi, L.
\newblock Spatio-temporal backpropagation for training high-performance spiking
  neural networks.
\newblock \emph{Frontiers in neuroscience}, 12:\penalty0 331, 2018.

\end{thebibliography}
\bibliographystyle{icml2021}

\newpage
\onecolumn
\appendix

\section{Conversion Error Analysis}
\label{sec_append_A}
We use $||\cdot||$ to denote the Frobenius norm. According to \autoref{eq_ann} and \autoref{eq_clip_floor}, the activation of ANN can be computed by
\begin{flalign}
    ||\mathbf{x}^{(n)}  - \bar{\mathbf{s}}^{(n)}|| &  = 
    ||h(\mathbf{W}^{(n-1)}\mathbf{x}^{(n-1)}) - g(\mathbf{W}^{(n-1)}\bar{\mathbf{s}}^{(n-1)})||  \nonumber\\
    & = ||h(\mathbf{W}^{(n-1)}\mathbf{x}^{(n-1)}) - h(\mathbf{W}^{(n-1)}\bar{\mathbf{s}}^{(n-1)}) + h(\mathbf{W}^{(n-1)}\bar{\mathbf{s}}^{(n-1)}) - g(\mathbf{W}^{(n-1)}\bar{\mathbf{s}}^{(n-1)})|| \\
    & \le ||\mathbf{W}^{(n-1)}(\mathbf{x}^{(n-1)} - \bar{\mathbf{s}}^{(n-1)}) + Err^{(n)}||, \label{eq_recur_err}
\end{flalign}
where $h(\cdot)$ is the $\mathrm{ReLU}$ function in ANN and $g(\cdot)$ is the $\mathrm{clipfloor}$ function in \autoref{eq_clip_floor}. \autoref{eq_recur_err} is based on the fact that activation $h(\cdot)$ is a piecewise linear function with gradient less than or equal to 1.
$Err^{(n)}=h(\mathbf{W}^{(n-1)}\bar{\mathbf{s}}^{(n-1)}) - g(\mathbf{W}^{(n-1)}\bar{\mathbf{s}}^{(n-1)})$. Recursively apply \autoref{eq_recur_err}, then we have
\begin{flalign}
    ||\mathbf{x}^{(n)} - \bar{\mathbf{s}}^{(n)}||
    & \le ||\mathbf{W}^{(n-1)}(\mathbf{x}^{(n-1)} - \bar{\mathbf{s}}^{(n-1)}) + Err^{(n)}|| \nonumber\\
    & \le ||\mathbf{W}^{(n-1)}\mathbf{W}^{(n-2)}(\mathbf{x}^{(n-2)} - \bar{\mathbf{s}}^{(n-2)}) + \mathbf{W}^{(n-1)}Err^{(n-1)} + Err^{(n)}|| \\
    & \le ||(\mathbf{x}^{(1)} - \bar{\mathbf{s}}^{(1)})\prod_{\ell=1}^n\mathbf{W}^{(\ell)} 
     + \sum_{\ell=1}^{n} Err^{(\ell)} \prod_{k=\ell}^{n}\mathbf{W}^{(k)}||
    \label{eq_final_err}
\end{flalign}
Note that the first term in \autoref{eq_final_err} is 0 if we use the same image input to ANN and SNN.

\section{Converting Average Pooling Layers}
Consider a batch of input to AvgPool2d layers with size of $[N,C,H,W]$ where $N$ is the batch size, $C$ is the channel number, $H$ and $W$ are the width and heights of inputs, as well as kernel size of AvgPool2d represented by $[kW, kH]$, we can precisely describe the AvgPool2d forwarding as
\begin{equation}
    out(N_i,C_j,h,w) = \frac{1}{kW\times kH} \sum_{m=0}^{kH-1}\sum_{n=0}^{kW-1}input(N_i, C_j, stride[0]\times h+m, stride[1]\times w+n).
\end{equation}
This forwarding function is a special case of depthwise Conv2d, where the kernel size of Conv2d is equal to $[kW, kH]$. The weights of the Conv2d should be constant for all elements in the kernel$\frac{1}{kW\times kH}$, and the bias of the Conv2d should be zero.

%For VGG-16~\cite{simonyan2014very}, there are 5 AvgPool2d layers in the ANN and the kernel size of them is $2\times2$. In that case, the AvgPool2d will produce outputs with 4 different possibilities, i.e. $[0,0.25,0.5,0.75,1]$. Thus the SNN forwarding cannot share the efficiency brought by binary activation.

We hereby conduct an ablation study that investigates the impact of AvgPool2d layers conversion. We test VGG-16 and ResNet-34 on ImageNet ($T=32$, Light Pipeline). VGG-16 has five $2\times 2$ AvgPool2d layers. ResNet-34 has one $2\times 2$ and one $7\times7$ AvgPool2d layers. On VGG-16, the accuracy is much lower if we convert the AvgPool2d layers. On ResNet-34, the impact of AvgPool2d layers is much lower than that on VGG-16, with only a 2\% accuracy drop if we convert AP.

\begin{table}[t]
    \footnotesize
        \centering
        \begin{adjustbox}{max width=0.48\textwidth}
        \begin{tabular}{lrr}
        \toprule
        \textbf{Method} & Convert AP  & Not Convert AP \\
        \midrule
        VGG-16 &  24.88 & 49.53 \\
        ResNet-34 & 50.21 & 52.87\\ 
        \bottomrule
        \end{tabular}
        \end{adjustbox}
        \caption{Impact of AvgPool2d layers in SNN.} \label{tab_avgpool}
    \vspace{-2mm}
\end{table}

\section{ANN Training Implementation}

\subsection{ImageNet}
The ImageNet dataset~\cite{deng2009imagenet} contains 120M training images and 50k validation images. For training pre-processing, we random crop and resize the training images to 224$\times$224. We additionally apply CollorJitter with brightness=0.2, contrast=0.2, saturation=0.2, and hue=0.1. For test images, they are center-cropped to the same size. For all architectures we tested, the Max Pooling layers are replaced to Average Pooling layers and are further converted to depthwise convolutional layers. The ResNet-34 contains a deep-stem layer (i.e., three 3$\times$3 conv. layers to replace the original 7$\times$7 first conv. layer) as described in~\citet{he2019bag}. We use Stochastic Gradients Descent with a momentum of 0.9 as the optimizer. The learning rate is set to 0.1 and followed by a cosine decay schedule~\cite{loshchilov2016sgdr}. Weight decay is set to $10^{-4}$, and the networks are optimized for 120 epochs. We also apply label smooth~\cite{szegedy2016labelsmooth}(factor=0.1) and EMA update with 0.999 decay rate to optimize the model.
For the MobileNet pre-trained model, we download it from \texttt{pytorchcv}\footnote{\url{https://pypi.org/project/pytorchcv/}}. 

\subsection{CIFAR}
The CIFAR 10 and CIFAR100 dataset~\cite{cifardataset} contains 50k training images and 10k validation images. We set padding to 4 and randomly cropped the training images to 32$\times$32. Other data augmentations include (1)random horizontal flip, (2) Cutout~\cite{devries2017cutout} and (3) AutoAugment~\cite{cubuk2019autoaugment}. For ResNet-20, we follow prior works~\cite{han2020rmp, han2020deep} who modify the official network structures proposed in~\citet{He2016Deep} to make a fair comparison. The modified ResNet-20 contains 4 stages with an additional deep-stem layer. For VGG-16 without BN layers, we add Dropout with a 0.25 drop rate to regularize the network. For MobileNet-CIFAR, we set the stride of the first conv. Layer to 1 to decreases the stage number to 4. For the model with BN layers, we use Stochastic Gradients Descent with a momentum of 0.9 as the optimizer. The learning rate is set to 0.1 and followed by a cosine decay schedule~\cite{loshchilov2016sgdr}. Weight decay is set to $5\times 10^{-4}$ and the networks are optimized for 300 epochs. For networks without BN layers, we set weight decay to $10^{-4}$ and learning rate to 0.005.

\section{Results on CIFAR}
\label{sec_cifar_results}
\begin{table*}[h]
    \footnotesize
        \centering
        \begin{adjustbox}{max width=\textwidth}
        \begin{tabular}{lcccccccc}
        \toprule
        \textbf{Method} & \textbf{Use BN} & \textbf{Convert AP} & \textbf{ANN Acc.} & \textbf{$T=32$} & \textbf{$T=64$} & \textbf{$T=128$} &\textbf{ $T=256$} & \textbf{ $T\ge 2048$} \\
        \midrule
        \multicolumn{9}{c}{\textbf{ResNet-20~\cite{He2016Deep}   CIFAR100}} \\
        \midrule
        Spike-Norm~\cite{Sengupta2018Going} & \xmark & \xmark & 69.72 & - &- &- &- & 64.09 \\ 
        RMP~\cite{han2020rmp} & \xmark & \xmark & 68.72 & 27.64 & 46.91 & 57.69 & 64.06 & 67.82 \\
        TSC~\cite{han2020deep} & \xmark & \xmark & 68.72 & - & - & 58.42 & 65.27 & 68.18 \\
        Opt.~\cite{deng2021optimal}* & \xmark & \cmark & 68.40 & 63.39 & 67.51 & 68.37 & 68.53 & 68.37 \\
        \textbf{Ours (Light Pipeline)} & \xmark & \cmark & 68.40 & \textbf{65.14} & \textbf{67.63} & \textbf{68.28} & \textbf{68.42} & 68.37 \\
        %\textbf{Ours (Advanced Pipeline)} & \xmark & \cmark & 68.40 & \textbf{-} & \textbf{-} & \textbf{-} & \textbf{-} & - \\
        Opt.~\cite{deng2021optimal}* & \cmark & \cmark & 77.16 & 51.27 & 70.12 & 75.81 & 77.22 & 77.19 \\
        \textbf{Ours (Light Pipeline)} & \cmark & \cmark & 77.16 & \textbf{75.53} & \textbf{77.08} & \textbf{77.50} & \textbf{77.59} &  77.25 \\
        \textbf{Ours (Advanced Pipeline)} & \cmark & \cmark & 77.16 & \textbf{76.32} & \textbf{77.29} & \textbf{77.73} & \textbf{77.63} & 77.25 \\
        \midrule
        \multicolumn{9}{c}{\textbf{VGG-16~\cite{simonyan2014very} CIFAR100}} \\
        \midrule
        Spike-Norm~\cite{Sengupta2018Going} & \xmark & \xmark & 71.22 & - &- &- &- & 70.77 \\ 
        RMP~\cite{han2020rmp} & \xmark & \xmark & 71.22 & - & - & 63.76 & 68.34 & 70.93 \\
        TSC~\cite{han2020deep} & \xmark & \xmark & 71.22 & - & - & 69.86 & 70.65 & 70.97 \\
        Opt.~\cite{deng2021optimal}* & \xmark & \cmark & 70.21 & 56.16 & 62.93 & 67.45 & 69.36 & 70.35 \\
        \textbf{Ours (Light Pipeline)} & \xmark & \cmark & 70.21 & \textbf{64.53} & \textbf{67.14} & \textbf{68.99} & \textbf{69.98} & 70.30 \\
        Opt.~\cite{deng2021optimal}* & \cmark & \cmark & 77.89 & 7.64 & 21.84 & 55.04 & 73.54 & 77.71\\
        \textbf{Ours (Light Pipeline)} & \cmark & \cmark & 77.89 & \textbf{65.73} & \textbf{72.38} & \textbf{75.82} & \textbf{77.12} & 77.87 \\    
        \textbf{Ours (Advanced Pipeline)} & \cmark & \cmark & 77.89 & \textbf{73.55} & \textbf{76.64} & \textbf{77.40} & \textbf{77.68} & 77.87 \\
        \midrule
        \multicolumn{9}{c}{\textbf{MobileNet~\cite{Howard2017MobileNetsEC} CIFAR100}} \\ 
        \midrule
        Opt.~\cite{deng2021optimal}* & \cmark & \cmark & 73.23 & 1.28 & 4.88  &39.39 & 65.79 & 73.01 \\    
        \textbf{Ours (Light Pipeline)} & \cmark & \cmark & 73.23 & \textbf{40.06} & \textbf{62.81} & \textbf{69.41}  & \textbf{71.98} & 73.19\\
        \textbf{Ours (Advanced Pipeline)} & \cmark & \cmark & 73.23 & \textbf{42.64} & \textbf{63.24} & \textbf{71.02} & \textbf{72.54} & 73.18\\
        \midrule
        \multicolumn{9}{c}{\textbf{ResNet-20~\cite{He2016Deep}   CIFAR10}} \\
        \midrule
        Spike-Norm~\cite{Sengupta2018Going} & \xmark & \xmark & 89.10 & - &- &- &- & 87.46 \\ 
        Hybrid Train~\cite{rathi2019enabling} & \xmark & \xmark & 93.15 & - &- &- & 92.22 & 92.94 \\
        RMP~\cite{han2020rmp} & \xmark & \xmark & 91.47 & - & - & 87.60 & 89.37 & 91.36 \\
        TSC~\cite{han2020deep} & \xmark & \xmark & 91.47 & - & 69.38 & 88.57 & 90.10 & 91.42 \\
        Opt.~\cite{deng2021optimal}* & \xmark & \cmark & 93.94 & 86.67 & 91.96 & 93.48 & 93.76 & 93.94 \\
        \textbf{Ours (Light Pipeline)} & \xmark & \cmark & 93.94 & \textbf{93.00} & \textbf{93.61} & \textbf{93.85} & \textbf{93.89} & 93.94 \\
       % \textbf{Ours (Advanced Pipeline)} & \xmark & \cmark & 93.94 & \textbf{92.71} & \textbf{93.56} & \textbf{93.83} & \textbf{-} & - \\
        Opt.~\cite{deng2021optimal}* & \cmark & \cmark & 95.46 & 84.06 & 92.48 & 94.68 & 95.30 & 94.42 \\
        \textbf{Ours (Light Pipeline)} & \cmark & \cmark & 95.46 & \textbf{94.44} & \textbf{95.20} & \textbf{95.29} & \textbf{95.36} &  95.47 \\
        \textbf{Ours (Advanced Pipeline)} & \cmark & \cmark & 95.46 & \textbf{94.78} & \textbf{95.30} & \textbf{95.42} & \textbf{95.41} & 95.45 \\
        \midrule
        \multicolumn{9}{c}{\textbf{VGG-16~\cite{simonyan2014very} CIFAR10}} \\
        \midrule
        Spike-Norm~\cite{Sengupta2018Going} & \xmark & \xmark & 91.70 & - &- &- &- & 91.55 \\ 
        Hybrid Train~\cite{rathi2019enabling} & \xmark & \xmark & 92.81 & - &- &91.13 & - & 92.48 \\
        RMP~\cite{han2020rmp} & \xmark & \xmark & 93.63 & 60.30 & 90.35 & 92.41 & 93.04 & 93.63 \\
        TSC~\cite{han2020deep} & \xmark & \xmark & 93.63 & - & 92.79 & 93.27 & 93.45 & 93.63 \\
        Opt.~\cite{deng2021optimal}* & \xmark & \cmark & 93.51 & 88.79 & 91.12 & 92.74 & 93.29 & 93.55 \\      
        \textbf{Ours (Light Pipeline)} & \xmark & \cmark & 93.51 & \textbf{91.82} & \textbf{92.57} & \textbf{93.21} & \textbf{93.35} & 93.54 \\
        Opt.~\cite{deng2021optimal}* & \cmark & \cmark & 95.72 & 76.24 & 90.64 & 94.11 & 95.33 & 95.73\\       
        \textbf{Ours (Light Pipeline)} & \cmark & \cmark & 95.72 &  \textbf{94.02} & \textbf{95.20} & \textbf{95.61} & \textbf{95.77} & 95.73 \\
        \textbf{Ours (Advanced Pipeline)} & \cmark & \cmark & 95.72 & \textbf{93.71} & \textbf{95.14}  & \textbf{95.65} & \textbf{95.79} & 95.79\\
        \midrule
        \multicolumn{9}{c}{\textbf{MobileNet~\cite{Howard2017MobileNetsEC} CIFAR10}} \\ 
        \midrule
        Opt.~\cite{deng2021optimal}* & \cmark & \cmark & 92.48 & 9.98 & 23.99  & 79.35 & 90.49 & 92.35 \\
        \textbf{Ours (Light Pipeline)} & \cmark & \cmark & 92.48 & \textbf{81.94} & \textbf{89.47}  & \textbf{91.61} & \textbf{92.20} &92.44\\
        \textbf{Ours (Advanced Pipeline)} & \cmark & \cmark & 92.48 & \textbf{82.37} & \textbf{90.40} & \textbf{91.70} & \textbf{92.29} & 92.47 \\ 
        \bottomrule
        \end{tabular}
        \end{adjustbox}
        \caption{Comparison of our algorithm with other existing SNN conversion works on CIFAR10 and CIFAR100. \textit{Use BN} means use BN layers to optimize ANN, \textit{Convert AP} means use Conv layers to replace Average Pooling layers. * denotes self-implementation results.} \label{tab_sota_compare}
    \vspace{-2mm}
\end{table*}

% \begin{table*}[h]
%     \footnotesize
%         \centering
%         \begin{adjustbox}{max width=\textwidth}
%         \begin{tabular}{lcccccccc}
%         \toprule
%         \textbf{Method} & \textbf{Use BN} & \textbf{Convert AP} & \textbf{ANN Acc.} & \textbf{$T=32$} & \textbf{$T=64$} & \textbf{$T=128$} &\textbf{ $T=256$} & \textbf{ $T\ge 2048$} \\

%         \bottomrule
%         \end{tabular}
%         \end{adjustbox}
%         \caption{Comparison of our algorithm with other existing SNN conversion works. \textit{Use BN} means use BN layers to optimize ANN, \textit{Convert AP} means use Conv layers to replace Average Pooling layers. * denotes self-implementation results.} \label{tab_sota_compare}
%     \vspace{-2mm}
% \end{table*}

\end{document}